\pdfoutput=1
\documentclass{article}
\usepackage{spconf,amsmath,amssymb,graphicx,url,times,booktabs,tabularx}
\usepackage{xspace}
\usepackage[english]{babel}
\usepackage{microtype}
\usepackage[hidelinks]{hyperref}
\let\originalthebibliography\thebibliography
\renewcommand{\thebibliography}[1]{%
  \originalthebibliography{#1}%
  \setlength{\itemsep}{3.4pt}%
  \setlength{\parsep}{0pt}%
  \setlength{\parskip}{0pt}%
}
\graphicspath{{figs/}}

\newcommand{\datasetname}{\mbox{AVPS-360}\xspace}

\title{AVPS-360: A Spatial Audio-Visual Benchmark for Instance Segmentation\\
       and Tracking of Sound-Producing Persons}

\name{Yifu Sun\sthanks{Corresponding author.}, Parthasaarathy Sudarsanam, Tuomas Virtanen, Archontis Politis}
\address{Audio Research Group, Tampere University, Tampere, Finland}

\begin{document}
\ninept
\let\small\normalsize
\let\footnotesize\normalsize
\maketitle

\begin{abstract}
Audio-visual instance segmentation (AVIS) segments and tracks individual
sound-producing objects in videos. Existing benchmarks predominantly use monaural
audio, which lacks inter-channel cues for spatially distinguishing concurrent
sound sources. We introduce
\datasetname{} (Audio-Visual Person Segmentation 360), a spatial audio-visual
benchmark bridging AVIS and sound event localisation and detection (SELD).
Built upon STARSS23, it pairs first-order Ambisonics audio with
synchronised $360^\circ$ video across 230 recordings totalling 10.5 hours.
Recordings average 164.0 seconds.
The dataset captures intermittent sound production in real-world interactions
dominated by speech, alongside laughter, clapping, and footsteps.
It provides 31{,}678 manually
reviewed instance masks linked to source identities and sound-event labels.
We also present a baseline that integrates spatial audio with visual features
for frame-level segmentation and identity association across frames.
Experiments on AVPS-360 show that existing AVIS methods struggle with
detection and identity association in long recordings with intermittent
sound production, while our baseline substantially improves performance in both
aspects.
\end{abstract}


\section{Introduction}
\label{sec:intro}

Vision and hearing help us associate sounds with the people around us and
follow their activities as they move and take turns speaking. For embodied
agents and assistive robots, similar awareness of the surrounding physical
environment could support interaction beyond a forward-facing camera's view,
such as responding to someone calling from behind or alerting users to
activity outside their current view.

\begin{table*}[!t]
\centering
\caption{Comparison of audio-visual segmentation datasets. \textbf{Len.} denotes
average recording duration; \textbf{\#Frames} counts annotated frames;
\textbf{Cls} denotes the number of object classes;
\textbf{Track} indicates persistent instance identities across
frames.}
\label{tab:datasets}
\vspace{4pt}
\setlength{\tabcolsep}{5pt}
\small
\begin{tabular}{@{}lcccrrrrrrc@{}}
\toprule
\textbf{Dataset} & \textbf{Domain} & \textbf{Video} & \textbf{Audio} & \textbf{\#Videos} & \textbf{Len.} & \textbf{Total} & \textbf{\#Frames} & \textbf{Masks} & \textbf{Cls} & \textbf{Track} \\
\midrule
AVSBench-O~\cite{zhou2022avs}      & Common & perspective & mono & 5{,}356  & 5.0\,s & 7.4\,h & 12{,}972 & 12{,}972 & 23 & $\times$ \\
AVSBench-S~\cite{zhou2023avss}     & Common & perspective & mono & 12{,}356 & 7.8\,s & 26.8\,h & 82{,}972 & -- & 70 & $\times$ \\
RefAVS-Bench~\cite{wang2024refavs} & Common & perspective & mono & 4{,}002  & 10.0\,s & 11.1\,h & 40{,}020 & --            & 51 & $\times$ \\
AVISeg~\cite{guo2024avis}          & Common & perspective & mono & 926      & 61.4\,s & 15.8\,h & 56{,}871 & 94{,}074      & 26 & \checkmark \\
AVTrack~\cite{wang2026avtrack}     & Human & perspective & mono\rlap{$^\dagger$} & 871 & 54.0\,s & 13.1\,h & 47{,}202\rlap{$^\ddagger$} & 36{,}216\rlap{$^\ddagger$} & 1 & \checkmark \\
\midrule
\datasetname{} (ours) & Human & panoramic & FOA & 230 & 164.0\,s & 10.5\,h & 37{,}718 & 31{,}678 & 1 & \checkmark \\
\bottomrule
\end{tabular}
\\[2pt]
{\footnotesize $^\dagger$ Audio is converted to mono by the published pipeline.
$^\ddagger$ Counts derived from the released data.}
\end{table*}

\begin{figure*}[!t]
\centering
\includegraphics[width=\textwidth]{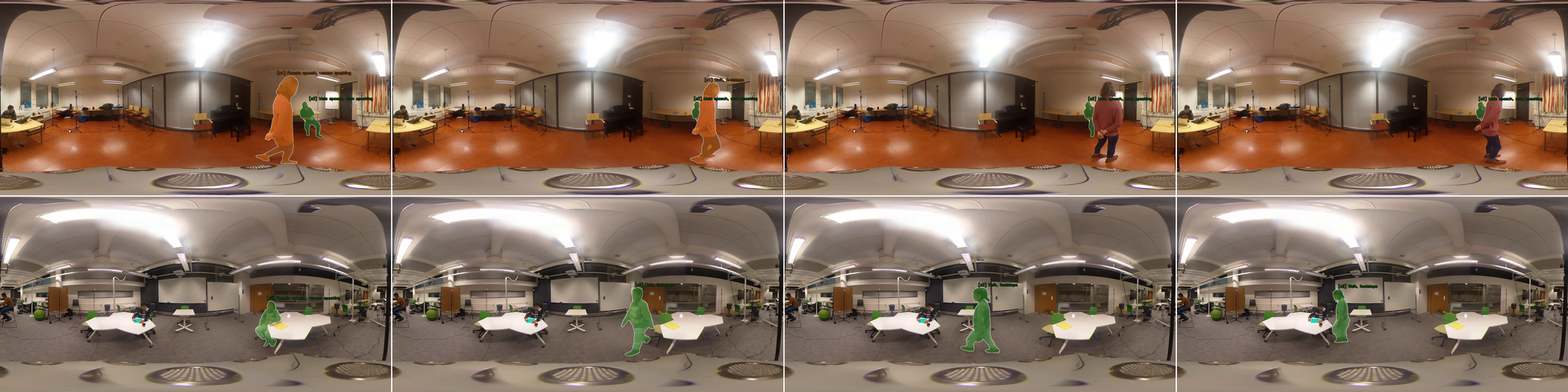}
\caption{Annotation examples from \datasetname{}. Coloured masks denote
sound-producing persons, illustrating occlusion (top) and sound-source movement
(bottom). Colours indicate consistent person identities within each recording.}
\label{fig:dataset_examples}
\end{figure*}

Audio-visual instance segmentation (AVIS) has established a foundation for
segmenting and tracking individual sound-producing objects through
AVISeg~\cite{guo2024avis}, while AVTrack~\cite{wang2026avtrack} extends evaluation
to complex human-centric scenes. However, these benchmarks primarily pair
perspective video with monaural audio and therefore do not jointly provide
panoramic visual context and inter-channel directional cues. General-purpose
audio-visual models such as Qwen3-Omni~\cite{qwen2025omni} likewise use monaural
audio in their standard input pipeline, without retaining inter-channel
directional cues.
Panoramic video
offers visual coverage beyond a single viewing direction, while spatial audio
provides directional evidence for determining which of several visible
persons are producing sound.

Audio-visual sound event localisation and detection (SELD) datasets such as
STARSS23~\cite{shimada2023starss23} provide multichannel audio and panoramic
video, but their direction-of-arrival (DOA) annotations specify source
directions rather than person masks. Standard SELD evaluation also does not
require persistent identities across overlapping or intermittent
events~\cite{diazguerra2024seld}. Linking spatial sound-event annotations to
person masks and recording-level identities would connect acoustic scene
analysis with instance-level visual grounding and tracking.

To connect these capabilities, we introduce \datasetname{}, built upon
STARSS23. It pairs four-channel first-order Ambisonics (FOA) audio with
synchronised $360^\circ$ video across 230 recordings of average duration
164.0 seconds, totalling 10.5 hours. We provide 31{,}678 manually reviewed person masks
linked to recording-level identities and sound-event labels. The indoor
interactions are dominated by speech, alongside laughter, clapping, and
footsteps. Models must segment visible persons while they produce sound,
suppress their masks during silence even when they remain visible, and
preserve their identities when sound production resumes.

We further establish a simple baseline that uses spatial audio to guide
frame-level instance segmentation and a memory of instance embeddings for
cross-frame identity association. Experiments demonstrate the benefits of
spatial audio and reveal substantial detection and association challenges for
existing AVIS methods on \datasetname{}.

Our contributions are: \textbf{(i)} a spatial audio-visual dataset linking
first-order Ambisonics and panoramic video to person masks, source identities,
and sound-event labels; \textbf{(ii)} a benchmark for segmenting and tracking
intermittently sound-producing persons in long recordings; and \textbf{(iii)} a
baseline and comparative evaluation demonstrating the value of spatial audio
and the challenges faced by existing AVIS methods.
The dataset is publicly available through the project
page\footnote{\url{https://stephensun.github.io/AVPS-360/}}. Code will be
released upon acceptance.

\section{Related Work}
\label{sec:related}

\noindent\textbf{Audio-visual segmentation and instance tracking.}
Visual sound localisation identifies audible sources in images through
response heatmaps~\cite{senocak2018learning,chen2021localizing}. Audio-visual
segmentation (AVS) predicts pixel-level masks,
with AVSBench supporting binary and semantic
segmentation~\cite{zhou2022avs,zhou2023avss} and RefAVS-Bench adding language-guided
targets~\cite{wang2024refavs}.
For instance-level modelling, AVISM~\cite{guo2024avis} combines frame-level
sound-source localisation with video-level object-token association.
Subsequent work investigates sound-source association~\cite{seo2026acvis} and
sound-enhanced online instance association~\cite{zhu2026seavis}.
SAMOT~\cite{peng2026samot} combines adaptive Mamba step modulation with
optimal transport for instance-level cross-modal matching.

\noindent\textbf{Sound event localisation and detection.}
SELD jointly identifies sound classes and estimates source directions from
multichannel audio~\cite{adavanne2018seld,politis2021overview}, including
simultaneously active sources of the same class~\cite{shimada2022multiaccdoa}.
Source locations are represented by azimuth and elevation.
Audio-visual SELD incorporates visual cues into event detection and
localisation through audio-visual embedding fusion~\cite{berghi2024fusion}
and video-attention mechanisms~\cite{hong2025mvanet}.

\section{The \datasetname{} Dataset}
\label{sec:dataset}

\datasetname{} is built on STARSS23~\cite{shimada2023starss23}, a dataset of
indoor sound scenes improvised by participants following general instructions
and recorded in 23 rooms with a spherical microphone array and a co-located
$360^\circ$ camera. It provides synchronised four-channel
first-order Ambisonics, equirectangular video at $1920\times960$ pixels and
29.97 frames per second, and spatiotemporal sound-event annotations.
Visible faces are blurred in the released STARSS23 videos for participant
privacy~\cite{shimada2023starss23}.
We keep the STARSS23 partition: its development set becomes our training split
and its evaluation set our test split, so no room or recording session is shared between
the two. A person track links the annotated masks of the same person throughout
one recording. The training split contains 156 recordings (7.06\,h), 20{,}734 masks,
and 340 person tracks; the test split contains 74 recordings (3.42\,h),
10{,}944 masks, and 168 person tracks. No separate validation split is used.

\begin{figure*}[t]
\centering
\includegraphics[width=0.95\textwidth]{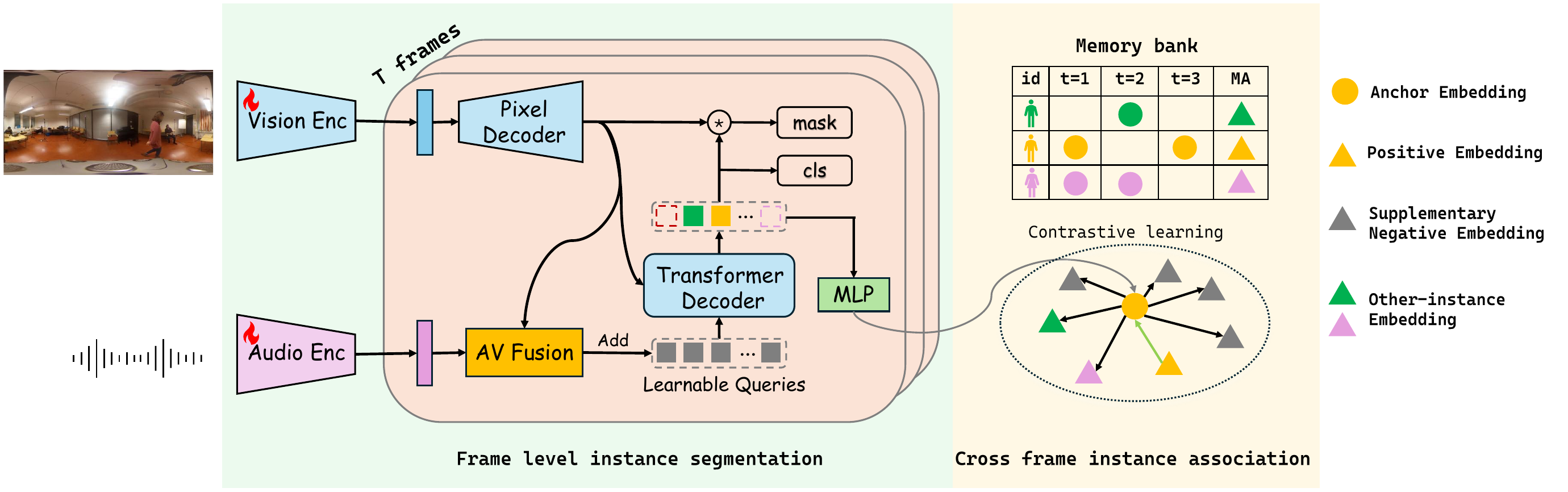}
\caption{Baseline architecture. Audio guides frame-level segmentation, and
the resulting instance queries are mapped to embeddings for cross-frame association. A memory bank
provides historical embeddings for contrastive learning. $T$ denotes the number
of sampled frames per training clip, and MA denotes momentum-averaged embeddings.}
\label{fig:model}
\end{figure*}

\subsection{Annotation}
We annotate one video frame per second, yielding 37{,}718 annotated frames.
Each visible sound-producing person receives one mask per timestamp and retains
the same identity across sound-event classes; simultaneous event labels are allowed.
Regions of the same person split across the panoramic seam are annotated as
a single instance mask.
For heavily occluded persons whose visible regions cannot be reliably annotated,
we retain sound-event labels and source identities but omit masks.
These persons are treated as non-targets at those timestamps.
Fig.~\ref{fig:dataset_examples} shows sample frames.

We use SAM~3~\cite{carion2025sam} to segment visible persons and retain five
human-produced sound classes from STARSS23: female speech, male speech,
clapping, laughter, and footsteps. Projected DOA locations are matched to the
nearest mask centroids to assign source identities and sound-event labels.
Non-human sounds remain as acoustic distractors. All annotated frames undergo
multiple rounds of manual review, correction, and verification using ISAT\footnote{\interfootnotelinepenalty=10000\relax
\url{https://github.com/yatengLG/ISAT_with_segment_anything}} to
resolve segmentation errors and identity mismatches caused by source-location
offsets, movement, and occlusion.
Event categories are provided as metadata but are neither used for event-class
supervision nor predicted by the baseline.
\subsection{Dataset statistics}
Table~\ref{tab:datasets} compares dataset scale, recording duration, sensing
modalities, and annotation coverage. \datasetname{} recordings average
164.0\,s, compared with 61.4\,s for AVISeg and 54.0\,s for AVTrack, providing
longer sequences for evaluating instance association. Its annotations comprise
508 recording-level person tracks, with persons appearing in different
recordings counted separately.

\begin{figure}[!tb]
\begin{minipage}[b]{0.49\columnwidth}
\centering
\includegraphics[width=\linewidth]{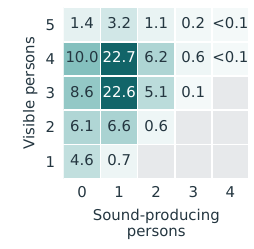}
\par (a)
\end{minipage}\hfill
\begin{minipage}[b]{0.49\columnwidth}
\centering
\includegraphics[width=\linewidth]{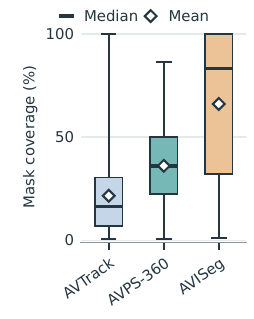}
\par (b)
\end{minipage}
\caption{Illustrations of \datasetname{} dataset statistics.}
\label{fig:dataset_statistics}
\end{figure}

Fig.~\ref{fig:dataset_statistics}(a) shows the joint distribution of
visible-person and sound-producing-person counts, with the latter counted from
nonempty annotated masks. Each cell reports a percentage of all annotated
frames in both splits. In 67.9\% of
frames, visible persons include both targets and non-targets; in 30.7\%, persons
are visible but none has a target mask. These cases illustrate target selection
among coexisting persons rather than segmentation of everyone visible.

Fig.~\ref{fig:dataset_statistics}(b) compares per-track temporal mask coverage,
defined as the proportion of a recording's annotated frames with a nonempty
mask for that track. Averaged over tracks across all splits, coverage is
36.1\% in \datasetname{}, 66.1\% in AVISeg~\cite{guo2024avis} (all object
categories), and 21.6\% in AVTrack~\cite{wang2026avtrack}. Boxes show
interquartile ranges and whiskers indicate minima and maxima. Coverage describes
annotated target presence rather than acoustic activity alone; missing masks
may reflect silence, severe occlusion, or absence from view.

\section{Baseline Model}
\label{sec:model}

To address long recordings with intermittent sound production, our baseline
combines audio-conditioned frame-level segmentation with memory-based identity
association (Fig.~\ref{fig:model}).

\subsection{Frame-level Instance Segmentation}
We build on Mask2Former~\cite{cheng2022mask2former}, using a visual backbone
and pixel decoder to extract multi-scale features $F_t^\ell$ and mask features
for frame $t$, with $\ell\in\{1,2,3\}$ indexing the visual scales.
We use the CNN--Conformer audio encoder from~\cite{berghi2024fusion},
pretrained for SELD on spatial sound scenes synthesised with the DCASE 2022
generator~\cite{politis2022starss22}.\footnote{\interfootnotelinepenalty=10000\relax
\url{https://github.com/danielkrause/DCASE2022-data-generator}}
It maps seven feature channels---four FOA log-mel spectrograms and
three cross-channel acoustic intensity-vector components---to an audio
embedding $a_t$ for each video frame $t$.

Following AVIS~\cite{guo2024avis}, a learned projection $W_b$ maps audio to
$b_t=W_ba_t\in\mathbb{R}^{D}$, where $D$ is the query feature dimension.
Let $N$ be the number of instance queries and
$P\in\mathbb{R}^{N\times D}$ their learnable positional embeddings.
Audio-visual fusion initialises the decoder queries as
\begin{equation}
\begin{aligned}
U_t^\ell &= \operatorname{CA}_{\ell}(b_t,F_t^\ell;P),\\
Q_t^0 &= Q_{\mathrm{learn}}+W_f[U_t^1;U_t^2;U_t^3],
\end{aligned}
\label{eq:frame_fusion}
\end{equation}
where each cross-attention block $\operatorname{CA}_{\ell}$ broadcasts $b_t$
to $N$ queries, adds $P$, and uses $F_t^\ell$ as keys and values, producing
$U_t^\ell\in\mathbb{R}^{N\times D}$. $Q_{\mathrm{learn}}\in\mathbb{R}^{N\times D}$
is learnable query content; brackets denote feature concatenation, and $W_f$
projects from $3D$ to $D$ dimensions.
The decoder attends to visual features to refine $Q_t^0$ into
$q_{t,i}\in\mathbb{R}^{D}$, $i\in\{1,\ldots,N\}$.
Classification and mask heads predict sound-producing-person labels and
instance masks, with Hungarian matching used during training.

\subsection{Cross-frame Instance Association}
Following CTVIS~\cite{ying2023ctvis}, we map audio-conditioned queries $q_{t,i}$
to association embeddings $e_{t,i}=\operatorname{MLP}(q_{t,i})\in\mathbb{R}^{D}$
using a three-layer multilayer perceptron (MLP). Contrastive learning aligns
same-person embeddings across frames and separates different identities.

The memory bank stores momentum-averaged embeddings for each identity.
For a current-frame anchor $e$, the contrastive loss is
\begin{equation}
\mathcal{L}_{\mathrm{emb}}(e)=-\log
\frac{\exp(e^{\top}e^{+})}
{\exp(e^{\top}e^{+})+\sum_{e^{-}\in\mathcal{N}_e}\exp(e^{\top}e^{-})},
\label{eq:contrastive}
\end{equation}
where $e^{+}$ is the same identity's memory embedding. The negative set
$\mathcal{N}_e$ contains other-instance memory embeddings and supplementary
negative embeddings sampled from other queries in the current frame.
Equation~\eqref{eq:contrastive} uses unnormalised embeddings without temperature
scaling. Averaging over eligible anchors gives
$\mathcal{L}_{\mathrm{emb}}$. The auxiliary loss $\mathcal{L}_{\mathrm{aux}}$
regresses anchor--positive and anchor--negative cosine similarities to one
and zero, respectively, using mean squared error.
The overall training objective is
\begin{equation}
\begin{aligned}
\mathcal{L}_{\mathrm{seg}} &= \lambda_{\mathrm{cls}}\mathcal{L}_{\mathrm{cls}}
    +\lambda_{\mathrm{bce}}\mathcal{L}_{\mathrm{bce}}
    +\lambda_{\mathrm{dice}}\mathcal{L}_{\mathrm{dice}},\\
\mathcal{L} &= \mathcal{L}_{\mathrm{seg}}
    +\lambda_{\mathrm{emb}}\mathcal{L}_{\mathrm{emb}}
    +\lambda_{\mathrm{aux}}\mathcal{L}_{\mathrm{aux}},
\end{aligned}
\label{eq:total_loss}
\end{equation}
where $\mathcal{L}_{\mathrm{cls}}$ is query cross-entropy against matched class
labels (no-object for unmatched queries), and $\mathcal{L}_{\mathrm{bce}}$ and
$\mathcal{L}_{\mathrm{dice}}$ are mask binary cross-entropy and Dice losses.
These segmentation losses and intermediate decoder supervision follow Mask2Former.
Loss weights follow CTVIS.

At inference, full recordings are processed sequentially at 1 fps, with
cosine similarity associating current embeddings with identities stored in memory.
Frame-wise confidence filtering selects output masks, while memory preserves
identities across temporary silence.

\section{Experiments and Results}
\label{sec:exp}

\noindent\textbf{Evaluation metrics.}
Following AVIS and AVTrack~\cite{guo2024avis,wang2026avtrack}, we report
Track AP, HOTA, DetA, AssA, and IDF1.
Track AP evaluates instance mask trajectories using spatio-temporal IoU.
HOTA~\cite{luiten2021trackeval} combines detection (DetA) and association (AssA),
while IDF1~\cite{ristani2016performance} measures identity consistency.

\noindent\textbf{Implementation details.}
We train our baseline for 24,000 iterations with AdamW
on two NVIDIA GH200 GPUs, using a batch of two clips with ten frames
sampled from a 20-second window per clip. The initial learning rate is
$10^{-4}$ ($10^{-5}$ for both encoders), reduced by a factor of 0.1 at
16,000 iterations.
During training, we apply clip-consistent panoramic shifts, horizontal flips,
and visual colour jitter. The FOA audio undergoes matching rotations and
reflections.

\subsection{Main Results}
Table~\ref{tab:main} compares our baseline with AVISM~\cite{guo2024avis}
and ACVIS~\cite{seo2026acvis}. SeaVIS~\cite{zhu2026seavis} and
SAMOT~\cite{peng2026samot} are not included because we could not find
public implementations during this study.
All compared models use visual weights pretrained on
ImageNet~\cite{deng2009imagenet} and COCO~\cite{lin2014coco}, and are trained
on the training split of \datasetname{}. The VGGish variants follow the audio pipelines of their
released implementations. For the spatial-audio variants, we replace
VGGish with the same pretrained CNN--Conformer used in our baseline and
fine-tune it jointly with the segmentation and tracking model, retaining
each method's fusion and tracking architecture.

Replacing VGGish with the CNN--Conformer increases AVISM's HOTA from
29.85 to 31.42, whereas ACVIS changes from 26.50 to 25.63. Our baseline
achieves 17.92 Track AP and 60.57 HOTA, outperforming all evaluated
configurations across the reported metrics (Table~\ref{tab:main}). We next
examine which advantages remain without audio and the additional benefit
of spatial cues.

\begin{table}[t]
\centering
\caption{Comparison of different AVIS methods on \datasetname{}.}
\label{tab:main}
\setlength{\tabcolsep}{2pt}
\footnotesize
\begin{tabular}{@{}llrrrrr@{}}
\toprule
Method & Audio Enc & AP & HOTA & DetA & AssA & IDF1 \\
\midrule
AVISM~\cite{guo2024avis} & VGGish     & 0.46  & 29.85 & 26.11 & 34.25 & 39.51 \\
AVISM~\cite{guo2024avis} & CNN--Conf. & 0.72  & 31.42 & 27.84 & 35.55 & 40.72 \\
ACVIS~\cite{seo2026acvis} & VGGish     & 0.68 & 26.50 & 19.98 & 35.29 & 31.42 \\
ACVIS~\cite{seo2026acvis} & CNN--Conf. & 1.58 & 25.63 & 19.59 & 33.67 & 29.87 \\
\midrule
Ours                    & CNN--Conf. & 17.92 & 60.57 & 59.34 & 62.00 & 73.43 \\
\bottomrule
\end{tabular}
\end{table}

\subsection{Ablation Studies}
\noindent\textbf{Audio input.}
Table~\ref{tab:fusion} compares a visual-only variant (None) with W-zero,
W-repeat, and FOA inputs. The visual-only variant bypasses the audio encoder and audio-visual fusion
while retaining the visual segmentation and association pipeline.
All audio variants share the CNN--Conformer
architecture and pretrained weights. W-zero and W-repeat use $[W,0,0,0]$
and $[W,W,W,W]$ waveforms during training and inference, with all seven
feature channels recomputed. Even without audio, our baseline exceeds all
evaluated AVISM/ACVIS variants in HOTA, DetA, AssA and IDF1
(Tables~\ref{tab:main}--\ref{tab:fusion}), although Track AP remains low
(0.51). This advantage is not attributable to spatial audio and is consistent
with a benefit of the frame-wise, memory-based design. FOA further improves
all metrics, raising Track AP from at most 2.30 among the controls to 17.92.

\begin{table}[t]
\centering
\caption{Audio input ablation.}
\label{tab:fusion}
\setlength{\tabcolsep}{2pt}
\footnotesize
\begin{tabular}{@{}lrrrrr@{}}
\toprule
Audio input & AP & HOTA & DetA & AssA & IDF1 \\
\midrule
None & 0.51 & 34.98 & 31.28 & 39.17 & 47.33 \\
W-zero & 0.29 & 34.16 & 30.85 & 37.88 & 46.61 \\
W-repeat & 2.30 & 42.99 & 40.56 & 45.68 & 57.69 \\
FOA & \textbf{17.92} & \textbf{60.57} & \textbf{59.34} & \textbf{62.00} & \textbf{73.43} \\
\bottomrule
\end{tabular}
\end{table}

\noindent\textbf{Visual backbone and pre-training.}
We examine the effects of the visual backbone and pre-training dataset while
keeping the spatial audio encoder unchanged (Table~\ref{tab:visual_init}).
ImageNet+COCO models are initialised from Mask2Former, while
ImageNet+COCO+OVIS uses CTVIS weights further fine-tuned on OVIS~\cite{qi2022ovis}.
With ResNet-50, OVIS pre-training improves Track AP and HOTA. Swin-L achieves
the best overall performance.

\begin{table}[!htbp]
\centering
\caption{Visual backbone and pre-training.}
\label{tab:visual_init}
\setlength{\tabcolsep}{2pt}
\footnotesize
\begin{tabular}{@{}llrrrrr@{}}
\toprule
Backbone & Initialisation & AP & HOTA & DetA & AssA & IDF1 \\
\midrule
ResNet-50 & IN+COCO & 17.92 & 60.57 & 59.34 & 62.00 & 73.43 \\
ResNet-50 & IN+COCO+OVIS & 19.03 & 61.14 & 60.08 & 62.38 & 72.00 \\
Swin-L & IN+COCO & \textbf{21.94} & \textbf{63.16} & \textbf{63.53} & \textbf{62.89} & \textbf{74.00} \\
\bottomrule
\end{tabular}
\end{table}

\section{Conclusion}
\label{sec:conclusion}

We introduced \datasetname{}, linking spatial audio and panoramic video to
sound-producing person masks, recording-level identities, and sound-event
labels. The benchmark supports segmentation and tracking during extended
interactions with intermittent target activity. Our baseline combines
audio-conditioned frame-level prediction with identity memory and outperforms
the evaluated AVIS methods on \datasetname{}. Under the same encoder
architecture and initialisation, FOA substantially outperforms W-only controls,
supporting the contribution of spatial acoustic information.

\section{Compliance with Ethical Standards}
This study adds annotations to publicly released STARSS23
recordings~\cite{shimada2023starss23}, without new participant recruitment
or data collection. The original study reports ethics approval and informed
consent for public release. We retain the face blurring in the released videos.

\clearpage
\bibliographystyle{IEEEbib}
\bibliography{refs}

\end{document}